%% file: paper.tex
\documentclass[11pt]{article}

\usepackage[table]{xcolor} %
\usepackage{multirow}      %
\usepackage[preprint]{configuration/acl}
\definecolor{LightGray}{gray}{0.92}
\definecolor{LightYellow}{RGB}{255, 253, 225}

\input{resources/preamble.tex}

\title{Do Thinking Tokens Help or Trap? \\ Towards More Efficient Large Reasoning Model}

\author{
    Bowen Ding\textsuperscript{\rm 1,4,\footnotemark[1]},
    Yuhan Chen\textsuperscript{\rm 2,\footnotemark[1]}, 
    Futing Wang\textsuperscript{\rm 1,4}, 
    Lingfeng Ming\textsuperscript{\rm 3}, and
    Tao Lin\textsuperscript{\rm 4,5,\footnotemark[2]}
    \\
    \textsuperscript{1} Zhejiang University \ 
    \textsuperscript{2} Boston University \
    \textsuperscript{3} ByteDance \\
    \textsuperscript{4} School of Engineering, Westlake University \\
    \textsuperscript{5} Research Center for Industries of the Future, Westlake University \\
    $^{4}$\texttt{\{dingbowen, wangfuting, lintao\}@westlake.edu.cn} \\
    $^{2}$\texttt{erv1n@bu.edu}, $^{3}$\texttt{minglingfeng@bytedance.com} \\
}

\begin{document}
\maketitle
\renewcommand{\thefootnote}{\fnsymbol{footnote}}
\footnotetext[1]{Contributed equally.}
\footnotetext[2]{Corresponding author.}
\renewcommand{\thefootnote}{\arabic{footnote}}

\input{resources/main}

\bibliography{resources/reference}

\appendix
\input{resources/appendix}

\end{document}

%% file: resources/preamble.tex
\usepackage{amsmath,amssymb,amsthm}

\newenvironment{talign*}
{\csname align*\endcsname}
{\endalign}

\usepackage[utf8]{inputenc}         %

\usepackage[T1]{fontenc}            %

\usepackage{url}                    %
\usepackage{booktabs}               %
\usepackage{amsfonts}               %
\usepackage{nicefrac}               %
\usepackage{xcolor}                 %
\usepackage{algorithm}
\usepackage{algorithmic}
\usepackage{graphicx}
\usepackage{subcaption}
\usepackage[flushleft]{threeparttable}
\usepackage{float}
\usepackage{multirow}
\usepackage{xspace}
\usepackage{natbib}
\usepackage{enumitem}
\usepackage[font=small]{caption}
\usepackage{autobreak}
\usepackage{sidecap}
\usepackage{wrapfig}
\usepackage{bbding}
\usepackage[toc, page, header]{appendix}
\usepackage{tikz}
\usepackage{pifont}
\usepackage{siunitx}       %
\usepackage{times}
\usepackage{latexsym}
\usepackage{mathtools}
\usepackage[most]{tcolorbox}
\usepackage{fontawesome}
\usepackage[normalem]{ulem}
\usepackage{lmodern}
\usepackage{mdframed}
\usepackage{adjustbox}

\usepackage{microtype}

\usepackage{inconsolata}

\hypersetup{
    colorlinks=true,
    linkcolor=blue,
    citecolor=blue,
    urlcolor=blue
}

\definecolor{coral}{RGB}{255,127,80}
\definecolor{darkgreen}{RGB}{0,100,0}
\definecolor{darkyellow}{RGB}{204,153,0}
\definecolor{salmon}{RGB}{250,128,114}
\definecolor{LightGray}{gray}{0.92}
\definecolor{LightYellow}{RGB}{255, 253, 225}

\usepackage{xspace}
\newcommand{\method}{\textsc{DuP-PO}\xspace}

%% file: resources/main.tex
\begin{abstract}\label{sec:abstract}
    Large Reasoning Models (LRMs) excel at solving complex problems but face an overthinking dilemma.
    When handling simple tasks, they often produce verbose responses overloaded with thinking tokens (e.g., \textit{wait, however}).
    These tokens trigger unnecessary high-level reasoning behaviors like reflection and backtracking, reducing efficiency.
    In this work, our pilot study reveals that these thinking-token-induced
    behaviors are not essential for effective problem-solving and may even hinder correct reasoning within
    constrained token budgets. We identify this phenomenon as the thinking trap. To mitigate this issue, we
    propose \textbf{Du}al \textbf{P}olicy \textbf{P}reference \textbf{O}ptimization (\method), a novel algorithm featuring:
    (1) A rollout sampling strategy that guarantees balanced exposure to responses with and without thinking tokens;
    (2) A fine-grained advantage control technique to dynamically regulate the prediction of target tokens;
    (3) A policy shaping method ensuring stable gradient contributions from thinking tokens.
    Experimental results on five popular math reasoning benchmarks show that \method performs well on the popular LRM,
    which significantly improves their token efficiency during reasoning, while achieving the superior
    performance of the base model.~\footnote{The project is still under progress and our codes will be released 
    at \url{https://github.com/Danield21/Dual-Policy-Preference-Optimization}}
\end{abstract}

\begin{figure}[t]
    \centering
    \includegraphics[width=1\linewidth]{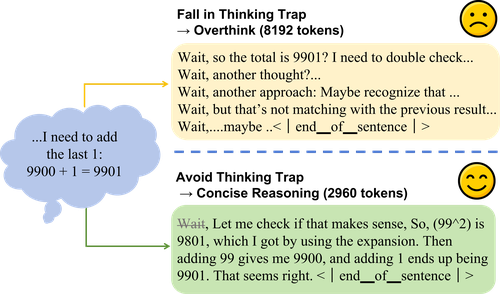}
    \caption{\textbf{The Illustration of Trapped v.s. Efficient Reasoning.}
        An example from \textsc{MATH500} where correct inference gets stuck in
        redundant verification loops, failing to produce a final answer within token
        limits.}
    \label{fig:thinking trap}
\end{figure}

\section{Introduction}
\label{sec:intro}
Large Reasoning Models (LRMs) excel at generating complex, human-like chains-of-thought (CoTs) for tasks like math, coding, and
 STEM Q\&A \citep{openai2024reasoning, deepseekai2025deepseekr1, qwen2025qwq, seed2025seed15thinking}.
Unlike conventional Large Language Models (LLMs), LRMs consistently generate discourse markers such as ``\textit{wait, hmm, however}'' within 
their reasoning processes, which we term \textbf{thinking tokens} (elaborated in Section~\ref{sec:implement_details}).
These tokens, in turn, activate advanced cognitive behaviours such as reflection, back-tracking, and thought transitions.
This phenomenon is often referred to as an ``aha moment'' and has traditionally been considered a hallmark of the evolution from a System-1 
agent to a System-2 agent \citep{deepseekai2025deepseekr1,zeng2025simplerlzooinvestigatingtamingzero,muennighoff2025s1}.

Recently, a growing number of studies~\citep{sui2025stopoverthinkingsurveyefficient,o1-pruner,feng2025efficientreasoningmodelssurvey} indicate 
that these advanced thought patterns in LRMs can lead to an explosive growth in response length.
Even for remarkably simple problems, such as ``$9900+1=?$'', LRM responses can become cluttered with extensive and unnecessary reflective thinking patterns, 
often resulting in outputs exceeding thousands of tokens, as shown in Figure~\ref{fig:thinking trap}.
This phenomenon, termed \textbf{overthinking}, significantly constrains the practical applicability of LRMs in real-world scenarios \citep{chen2025do}.

Existing training-based approaches tackle overthinking by collecting variable-length CoT~\citep{xia2025tokenskip,kang2025c3ot,ma2025cot,munkhbat2025self} and 
explicitly penalizing verbose responses \citep{fu2024efficiently,yu2024distilling,aggarwal2025l1}.
Nevertheless, these methods struggle to achieve an optimal balance between performance gains and token efficiency \citep{thinkless,thinkingfastrightbalancing}, highlighting 
the need for a deeper understanding of overthinking mechanisms to develop more effective solutions.

Recent work by \citet{muennighoff2025s1simpletesttimescaling} demonstrates that appending thinking tokens (e.g., \textit{wait}) after end-of-thinking delimiters can artificially 
extend reasoning duration, suggesting that frequent sampling of such tokens may contribute to overthinking.
However, the research community has yet to reach consensus on the role that thinking tokens play in the reasoning process, leading us to pose a fundamental question:
\begin{center}
    \textit{Do Thinking Tokens Help or Trap?}
\end{center}

To answer this question, we conduct systematic experiments analyzing thinking token behaviors in LRMs.
Our analysis reveals that thinking tokens may trigger a \textbf{thinking trap}, where unproductive reasoning loops that waste computational resources without improving task performance.
This challenges the prevailing assumption that more thinking necessarily leads to better reasoning.

Building on the insight, we propose a simple yet effective \textbf{r}einforcement \textbf{l}earning (RL) algorithm: \textbf{Du}al-\textbf{P}olicy \textbf{P}reference \textbf{O}ptimization (\method), which 
features three key innovations:
(1) \textbf{Dual-Policy Sampling} that provides balanced exposure to responses with and without thinking tokens during training;
(2) \textbf{Token-Level Advantage Scaling} that finely controls the reinforcement of specific tokens based on their utility;
and (3) \textbf{Policy Shaping} that ensures stable gradient flow for thinking token suppression.
This approach enables models to learn when thinking tokens are beneficial versus detrimental, avoiding the thinking trap while maintaining reasoning quality.
The main contributions of this work include:
\begin{itemize}[nosep, leftmargin=12pt]
    \item We identify and characterize the \textbf{thinking trap}, where thinking tokens drive unproductive reasoning cycles, providing empirical evidence that challenges their assumed necessity.
    \item We propose \textbf{\method}, a training-based approach that achieves superior performance-efficiency trade-offs by reinforcing concise correct responses while suppressing problematic 
    thinking token generation.
    \item We demonstrate \method's effectiveness across diverse mathematical reasoning benchmarks, achieving 6-20\% token reduction with performance improvements through minimal 
    training overhead.
\end{itemize}

\section{Related work}\label{sec:related_work}
\subsection{Analysis on Thinking Tokens} \label{sec:analysis_on_thining_tokens}
Since the advent of DeepSeek-R1, thinking tokens, predominantly featuring \textit{wait}, have garnered significant attention.
\citet{deepseekai2025deepseekr1} emphasizes the significance of these tokens, positing that their emergence signals a model's autonomous development of advanced problem-solving strategies.
Concurrently, \citet{zhou2025r1zero} and \citet{huggingface2025mini} have regarded the model's exhibition of ``aha moments'' deduced by thinking tokens as an indicator of successfully 
replicating R1.
Furthermore, other works have investigated the origins of these thinking tokens.
\citet{liu2025there} suggest that thinking tokens represent pre-existing patterns within base models, which RL methods merely activate.
\looseness=-1

Moreover, certain studies have delved into the model behaviors and internal mechanisms induced by thinking tokens.
\citet{yang2025understanding} posit that these thinking tokens can subtly alter the model's perception of problem difficulty, thereby facilitating the resolution of complex problems.
\citet{wang2025beyond} observe that in Qwen3 series models, tokens like \textit{wait} introduce high uncertainty.
Additionally, \citet{qian2025demystifying}, by calculating the mutual information between response tokens and the final answer, find mutual information peaks associated with thinking tokens such 
as \textit{hmm}, \textit{wait}, and \textit{therefore}, suggesting that these tokens possess superior representational capabilities for the answer compared to other tokens.

In contrast, our analysis demonstrates that thinking tokens can induce a trap of redundant cyclic verification, leading to reasoning inefficiencies such as overthinking in the context of mathematical 
reasoning tasks.

\subsection{Mitigating Overthinking}\label{sec:mitigate_overthinking}
To reduce over-thinking in LRMs, prior work explores both training-free and training-based approaches. Training-free methods include prompt-based techniques, which elicit concise reasoning through carefully designed prompts~\citep{ma2025reasoning}, and decoding-based strategies that terminate reasoning based on uncertainty signals or penalize unstable token transitions~\citep{fu2024efficiently,yang2025dynamic,wang2025thoughts}.

Training-based methods promote efficiency through supervised fine-tuning on variable-length CoT data~\citep{xia2025tokenskip,kang2025c3ot,ma2025cot,munkhbat2025self,yu2024distilling,cui2025stepwise} or reinforcement learning with length-aware rewards to discourage verbosity~\citep{yeo2025demystifying,luo2025o1,team2025kimi,aggarwal2025l1,shen2025dast,qu2025optimizing,hou2025thinkprune,yang2025think}.
Hybrid approaches further adapt reasoning depth based on task complexity~\citep{zhang2025adaptthink,tu2025learning}.

In contrast, our method \method requires no curated CoT data or predefined reasoning length.
It finely regulates thinking token usage, delivering a favorable balance between performance and efficiency.

\begin{figure*}[t]
    \centering
    \begin{subfigure}[b]{0.3\textwidth}
        \centering
        \includegraphics[width=1\textwidth, height=0.7\textwidth, keepaspectratio=false]{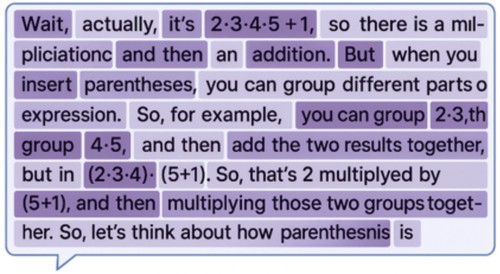}
        \caption{\small The heatmap of token probability. }
        \label{fig:prob}

    \end{subfigure}\hfill
    \begin{subfigure}[b]{0.33\textwidth}
        \centering
        \includegraphics[width=\textwidth]{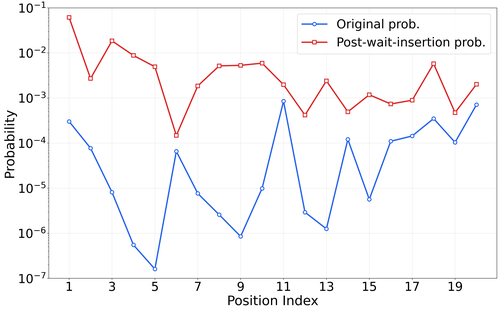}
        \caption{\small The illustration of \textit{wait} cascading effect.}
        \label{fig:insertion}

    \end{subfigure}\hfill
    \begin{subfigure}[b]{0.33\textwidth}
        \centering
        \includegraphics[width=\textwidth]{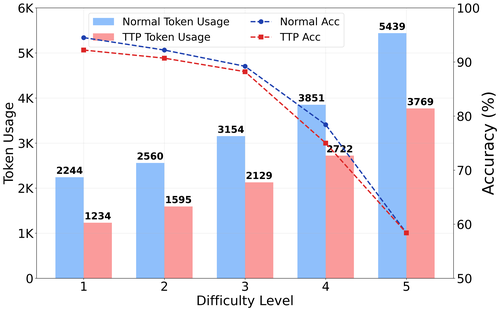}
        \caption{\small Normal reasoning v.s. ThinkTokenPenalty (TTP) on \textsc{MATH500}. }
        \label{fig:math500_accuracy}

    \end{subfigure}
    \caption{\textbf{Analysis of thinking token effects in DeepSeek-R1-Distill-Qwen-1.5B}.
        (\ref{fig:prob}) \textbf{The heatmap of token probability.}
        Darker backgrounds indicate higher probabilities, showing systematic over-confidence in thinking tokens like \emph{wait}.
        (\ref{fig:insertion}) \textbf{The illustration of \textit{wait} cascading effect.}
        Blue circles show the original \emph{wait} probabilities, red squares show $100$-fold increases after inserting a single \emph{wait} token, demonstrating auto-regressive amplification across $20$ positions.
        (\ref{fig:math500_accuracy}) \textbf{Normal reasoning v.s. ThinkTokenPenalty (TTP) on MATH500.}
        The performance and response token usuage across difficulty levels 1-5 is shown, where bars show token usage and dotted lines show accuracy for normal reasoning (blue) and TTP (red).
        TTP maintains comparable accuracy while reducing token consumption.
    }
    \label{fig:three_images}
\end{figure*}

\section{Rethinking the Role of Thinking Tokens}\label{sec:rethink_role} 
In this section, we challenge the prevailing assumption that thinking tokens universally enhance reasoning capabilities in large reasoning 
models~\citep{deepseekai2025deepseekr1,muennighoff2025s1}.

Our empirical analysis of 6,023 test responses from DeepSeek-R1-Distill-Qwen-1.5B~\cite{deepseekai2025deepseekr1}
reveals a counterintuitive pattern:
\begin{tcolorbox}[colback=blue!10,colframe=blue!50!black,title=\textbf{Observation}]\label{obs:think_token_freq}
    Incorrect responses contain \textbf{twice as many thinking tokens} as correct responses.
\end{tcolorbox}
This finding suggests that thinking token density correlates more strongly with reasoning failures than successes,
motivating a fundamental question:
\begin{tcolorbox}[colback=red!10,colframe=red!50!black,title=\textbf{Question}]
    Do thinking tokens \textbf{genuinely facilitate complex reasoning}? or do they \textbf{introduce detrimental overthinking}?
\end{tcolorbox}

We address this question through two complementary investigations:
Section~\ref{sec:think_token_absence_maintain_perf} examines whether reasoning performance degrades when thinking tokens are removed, while 
Section~\ref{sec:think_token_property} analyzes specific properties of thinking tokens that may contribute to overthinking behaviors in large reasoning models.

\subsection{When Fewer Thinking Tokens Maintain a Good Reasoning}\label{sec:think_token_absence_maintain_perf} %
To directly test the necessity of thinking tokens, we conduct a controlled experiment where thinking token generation is systematically suppressed through logits 
penalty (i.e., ThinkTokenPenalty in Section~\ref{sec:baselines}) on common thinking tokens (\textit{wait, hmm, hold on, alternatively, maybe, however, but, okay}),
Then we get a counterintuitive result:
\begin{tcolorbox}[colback=blue!10,colframe=blue!50!black,title=\textbf{Observation}]
    Suppressing thinking token generation produces \textbf{minimal degradation} in reasoning performance across difficulty levels.
\end{tcolorbox}

As illustrated in Figure~\ref{fig:math500_accuracy}, accuracy remains remarkably stable across \textsc{MATH500} difficulty levels, 
with both normal reasoning and ThinkTokenPenalty 
achieving comparable performance across varying problem difficulties.
Simultaneously, the bar chart reveals substantial savings on the reasoning cost, achieving approximately $1{,}000$ token reductions per 
response across different difficulty levels through 
disenabling thinking tokens.
Furthermore, our error analysis demonstrates that ThinkTokenPenalty significantly reduces reasoning failures attributed to excessive reflection 
and response truncation, decreasing such 
failures from $86\%$ to $37\%$.
These empirical findings lead to a critical takeaway:
\begin{tcolorbox}[colback=green!10,colframe=green!50!black,title=\textbf{Takeaway}]
    Thinking tokens are \textbf{not necessary} for effective reasoning;
    their absence often \textbf{improves} reasoning efficiency.
\end{tcolorbox}

We formalize this counterproductive phenomenon as \textbf{thinking trap}-reasoning deadlocks
caused by excessive self-reflection, as exemplified in Figure~\ref{fig:thinking trap}.

\subsection{Mechanisms Underlying Thinking Traps}\label{sec:think_token_property}
Having established that thinking tokens can impair reasoning performance, we investigate the underlying mechanisms driving this phenomenon.
Our analysis identifies two critical properties that enable thinking trap formation:
\textbf{over-confidence} and \textbf{cascading generation}.

\paragraph*{Over-confidence in Thinking Token Prediction.}
Analysis of our $6{,}023$ response dataset reveals an average of $106$ thinking tokens per response, indicating remarkably 
high generation frequency.
Moreover, Figure~\ref{fig:prob} demonstrates consistent high-probability patterns for thinking tokens such as \textit{wait} and \textit{but}, 
evidenced by the prominent highlighting behavior.
These observations suggest:
\begin{tcolorbox}[colback=yellow!10,colframe=yellow!50!black,title=\textbf{Hypothesis}]
    Models assign consistently \textbf{high probabilities} to thinking tokens, leading to their \textbf{frequent sampling}.
\end{tcolorbox}

Among these, \textit{wait} emerges as the most prevalent thinking token, constituting $37\%$ of all thinking tokens across responses on average.
We therefore focus on \textit{wait} tokens to validate our hypothesis.
To test this, we analyze the top $100$ responses with the highest \textit{wait} token occurrences, encompassing $492$ instances in total.
The average model-predicted probability for \textit{wait} at these positions reaches $0.88$, confirming
systematic over-confidence in \textit{wait} token generation within the LRM.

\paragraph*{Cascading Generation Patterns.}\label{sec:chain}
Further analysis of \textit{wait} tokens reveals another concerning pattern:
$96\%$ of responses contain multiple (more than one) \textit{wait} tokens.
Considering the auto-regressive mechanism of LRM generation, we hypothesize:
\begin{tcolorbox}[colback=yellow!10,colframe=yellow!50!black,title=\textbf{Hypothesis}]
    Previous thinking token generation \textbf{triggers cascading production} of additional thinking tokens.
\end{tcolorbox}
\looseness=-1

We validate this through controlled insertion experiments on $100$ responses initially containing no \textit{wait}
tokens (typically under $3{,}000$ tokens).
Inserting a single \textit{wait} token increases subsequent \textit{wait} prediction probabilities by $100$-fold over the following $20$ positions (Figure~\ref{fig:insertion}), 
demonstrating strong auto-regressive amplification.

Together, these findings reveal the key mechanisms underlying thinking trap formation:
\begin{tcolorbox}[colback=green!10,colframe=green!50!black,title=\textbf{Takeaway}]
    LRMs exhibit \textbf{systematic over-confidence} in thinking token utility and \textbf{cascading generation behaviors} that \textbf{create unproductive reasoning loops}.
\end{tcolorbox}

\section{Methodology}\label{sec:methodology}
Given the insights gained from our analysis on thinking tokens and their detrimental effects on reasoning performance, we propose \textbf{Du}al-\textbf{P}olicy 
\textbf{P}reference \textbf{O}ptimization 
(\method) to mitigate the thinking trap dilemma.

\method extends GRPO to \textbf{precisely control thinking token usage}.
Rather than eliminating thinking tokens arbitrarily, \method learns to strategically regulate their generation, enabling models to engage in productive reasoning while 
avoiding the thinking trap.

In this section, we first provide an overview of GRPO in Section~\ref{sec:preliminary}, followed by a detailed
explanation of the core concepts and implementation of \method in Section~\ref{sec:duppo_2}.

\subsection{Preliminary}\label{sec:preliminary}
\looseness=-1
In this section, we review the key components of GRPO that underpin our approach.

\subsection{GRPO}\label{sec:grpo}
The GRPO algorithm streamlines PPO~\citep{ppo} by replacing the traditional value network with group-based advantage estimation.
The algorithm operates by sampling $G$ response trajectories $\{\boldsymbol{\tau}_i\}_{i=1}^{G}$ from the current policy $\pi_{\theta_{\text{old}}}$ for each query-answer 
pair $(\boldsymbol{q}, \boldsymbol{a})$ in dataset $\mathcal{D}$.
Each trajectory receives a reward score $R_i$ through rule-based evaluation.

The binary reward function exemplifies this approach:
\begin{equation}
    \label{eq:binary_reward}
    R_i =
    \begin{cases}
        1.0 & \quad \text{if } \texttt{is\_equivalent}(\boldsymbol{a}, \boldsymbol{\tau}_i) \\
        0.0 & \quad \text{otherwise}
    \end{cases}
\end{equation}

This formulation creates a clear preference hierarchy: trajectories satisfying $\texttt{is\_equivalent}(\boldsymbol{a}, \boldsymbol{\tau}_i)$ are designated as \textbf{preferred} and receive positive 
rewards, while all other trajectories are treated as \textbf{unpreferred} with zero rewards.

\paragraph{Advantage Estimation.}\label{para:adv_estimation}
With these reward assignments, GRPO transforms the group-level preferences into token-level training signals.
The advantage for any token $\boldsymbol{\tau}_{i,t}$ is computed through group normalization:
\begin{equation}
    \label{eq:grpo_adv}
    A_{i}^{t} = \frac{R_i - \text{mean}(\{R_i\}_{i=1}^{G})}{\text{std}(\{R_i\}_{i=1}^{G})}
\end{equation}
This equation indicates that GRPO assigns the identical advantage to all tokens within a trajectory, treating all reasoning content identically regardless of their distinct roles in response quality.

\paragraph{Policy Optimization Objective.}
GRPO aims to maximize the expected advantage of the policy while preventing the current policy from deviating excessively from the reference policy.
The optimization objective $\mathcal{J}(\theta)$ incorporates token-level advantages through a clipped policy gradient formulation that balances preference learning with training stability:

\begin{small}
    \begin{align}
        \label{eq:grpo_obj}
        \mathcal{J}(\theta) & = \mathbb{E}_{\mathcal{D},\pi_{\theta_{\text{old}}}} \left[ \frac{1}{\sum_{i=1}^{G} |\tau_{i}|} \sum_{i=1}^{G} \sum_{t=1}^{|\tau_{i}|} \min \left( r_{i}^{t} A_{i}^{t}, C_{i}^{t} A_{i}^{t} \right) \right] \\
                            & \quad - \beta \cdot \mathbb{D}_{\text{KL}} \left[ \pi_{\theta} ||\pi_{\text{ref}} \right] \notag
    \end{align}
\end{small}%

where $r_{i}^{t}=\frac{\pi_{\theta}(\tau_{i,t}|\boldsymbol{q},\tau_{i,<t})}{\pi_{\theta_{\text{old}}}(\tau_{i,t}|\boldsymbol{q},\tau_{i,<t})}$ is the importance ratio measuring probability shifts for each token, and 
$C_{i}^{t} = \text{clip} (r_{i}^{t}, 1-\epsilon, 1+\epsilon)$ applies clipping to maintain training stability.
The clipping trick ensures that when the importance ratio exceeds the trust region bounds $[1-\epsilon, 1+\epsilon]$, the gradient contribution is capped, preventing excessive policy updates that could destabilize training.
\looseness=-1

\subsection{Token-Level Policy Gradient}\label{sec:policy_gradient_grpo}
As a policy gradient method, GRPO utilizes the policy gradient to control the token prediction. Building on the optimization objective in
Equation~\eqref{eq:grpo_obj}, the policy gradient for token $\boldsymbol{\tau}_{i,t}$
follows a conditional update rule:
\looseness=-1

\begin{small}
    \begin{equation}
        \textstyle
        \label{eq:policy_gradient}
        \nabla_{\theta} \mathcal{J}(\boldsymbol{\tau}_{i,t}) =
        \begin{cases}
            \nabla_{\theta} \log \pi_{\theta}(\boldsymbol{\tau}_{i,t}) A_{i}^{t} & \text{if } A_{i}^{t} > 0           \\
                                                                                 & \text{and } r_{i}^{t} < 1+\epsilon \\[1ex]
            \nabla_{\theta} \log \pi_{\theta}(\boldsymbol{\tau}_{i,t}) A_{i}^{t} & \text{if } A_{i}^{t} < 0           \\
                                                                                 & \text{and } r_{i}^{t} > 1-\epsilon \\[1ex]
            0,                                                                   & \text{otherwise}
        \end{cases}
    \end{equation}
\end{small}%
which exposes three fundamental properties:
\begin{itemize}[leftmargin=12pt, nosep]
    \item \textbf{Magnitude Scaling:}
          The advantage magnitude $|A_{i}^{t}|$ directly controls gradient strength, allowing proportional reinforcement based on trajectory success.
          Higher-performing trajectories generate stronger learning signals for all constituent tokens.

    \item \textbf{Directional Control:}
          The advantage sign determines whether token probabilities increase (positive advantage from preferred trajectories) or decrease (negative advantage from un-preferred ones), providing clear directional guidance for policy updates.

    \item \textbf{Gradient Gating:}
          The importance ratio $r_{i}^{t}$ acts as a gate, zeroing gradients when policy changes exceed clipping bounds.
          This mechanism prevents unstable updates while preserving meaningful learning signals within the trust region.
\end{itemize}

These properties form the foundation for \method's ability to apply distinct training signals to some concerned tokens (i.e., thinking tokens) versus others, enabling precise control over when and how models engage in self-reflection.
We leverage these insights to design our algorithm in Section~\ref{sec:duppo_2}.

\subsection{\method}\label{sec:duppo_2}
\method addresses two key limitations of GRPO: its inability to preferentially reinforce concise preferred trajectories and its failure to effectively suppress unpreferred trajectories that fall into the thinking trap.

The core training objective of \method extends GRPO's formulation:
\begin{equation}
    \label{eq:duppo-obj}
    \begin{aligned}
         & \mathbb{E}_{\mathcal{D}, (\pi_n, \pi_r)} \Big[ \frac{1}{\sum_{i=1}^{N+M} |\boldsymbol{\tau}_i|} \sum_{i=1}^{N+M} \sum_{t=1}^{|\boldsymbol{\tau}_i|} \\
         & \, \min \big( \hat{r}_i^t \hat{A}_i^t, \text{clip} (\hat{r}_i^t, 1-\epsilon, 1+\epsilon) \hat{A}_i^t \big) \Big]                                    \\
         & \, - \beta \cdot \mathbb{D}_{\text{KL}} \left[ \pi_{\theta} ||\pi_{\text{ref}} \right] \,.
    \end{aligned}
\end{equation}
where $\hat{A}_i^t = m_i^t \cdot A_i^t$ represents scaled advantages and $\hat{r}_i^t$ denotes calibrated importance ratios.
\looseness=-1
It enables three complementary components: \looseness=-1
\begin{enumerate}[nosep, leftmargin=12pt]
    \item \textbf{Dual-Policy Sampling} generates both thinking-heavy and thinking-free trajectories during rollout, providing comparative examples that GRPO's single-policy sampling cannot achieve.
    \item \textbf{Token-level Advantage Scaling} breaks GRPO's uniform advantage constraint by applying differential scaling factors $m_i^t$ to thinking tokens based on the trajectory's characteristic.
    \item \textbf{Policy Shaping} ensures a consistent gradient flow for thinking token suppression by calibrating importance ratios $\hat{r}_i^t$, overcoming GRPO's clipping limitations that can zero out crucial learning signals.
\end{enumerate}

\subsubsection{Dual-Policy Sampling}\label{sec:dual_policy_sampling}
Our first innovation addresses the single-policy sampling limitation of GRPO through a dual-policy approach during the rollout phase.

\paragraph{GRPO Sampling Limitation.}
GRPO samples all trajectories from a single policy $\pi_{\theta_{\text{old}}}$ (referred to as the normal policy $\pi_n$).
For models prone to overthinking, this creates a critical problem: nearly all sampled trajectories contain excessive thinking tokens, providing no examples of the concise responses we want to encourage.
Without seeing both thinking-heavy and thinking-free responses for the same query, GRPO cannot learn to prefer concise reasoning over overthinking.

\paragraph{Rectified Policy Design.}
To provide balanced training examples, we introduce a rectified policy ($\pi_{r}$) that systematically eliminates thinking tokens during generation.
This policy operates by setting the logit values of predefined thinking tokens to $-\infty$, effectively zeroing their generation probability.
For any token $\boldsymbol{\tau}_{\cdot,t} \in \mathcal{S}_{\text{think}}$ (our predefined set of thinking tokens), the rectified policy is defined as:

\begin{small}
    \begin{equation*}
        \label{eq:rectified_policy}
        \pi_{r}(\boldsymbol{\tau}_{\cdot,t} | \boldsymbol{q}, \boldsymbol{\tau}_{\cdot,<t}) =
        \begin{cases}
            \delta \approx 0,                                                                    & \text{if } \boldsymbol{\tau}_{\cdot,t} \in \mathcal{S}_{\text{think}}    \\
            \pi_{n}(\boldsymbol{\tau}_{\cdot,t} | \boldsymbol{q}, \boldsymbol{\tau}_{\cdot,<t}), & \text{if } \boldsymbol{\tau}_{\cdot,t} \notin \mathcal{S}_{\text{think}}
        \end{cases}
    \end{equation*}
\end{small}

\paragraph{Balanced Trajectory Generation.}
During rollout, \method samples $N$ trajectories from the rectified policy $\{\boldsymbol{\tau}^{r}_{i} \}_{i=1}^{N} \sim \pi_{r}$ and $M$ trajectories from the normal policy $\{\boldsymbol{\tau}^{n}_{i} \}_{i=1}^{M} \sim \pi_{n}$.

This dual-sampling strategy ensures the model observes both response types for each query, enabling it to learn the comparative value of concise versus thinking-heavy approaches.
Unlike GRPO's uniform sampling, this provides the contrastive examples necessary for effective thinking token regulation.

\subsubsection{Token-Level Advantage Scaling} \label{sec: token_level_advantage_scaling}
Our second innovation breaks GRPO's constraint of assigning the identical advantage within a trajectory.
\method applies token-specific advantage scaling based on the token's attribute and the corresponding trajectory preference.

\paragraph{GRPO Identical Advantage Limitation.}
Section~\ref{sec:grpo} demonstrates that GRPO assigns identical advantages to all tokens within a trajectory, limiting fine-grained control over token-level optimization.
Since advantage magnitude and sign directly determine each token's likelihood adjustment in the updated policy (see Section~\ref{sec:policy_gradient_grpo}), this uniform treatment creates a fundamental constraint:
GRPO cannot selectively encourage concise reasoning while suppressing certain overthinking triggers (i.e., thinking tokens) within the same trajectory.

\paragraph{Advantage Scaling Mechanism.}
To overcome this limitation, we introduce a scaling factor $m_{i}^{t}$ that modifies the original advantage $A_{i}^{t}$ to produce calibrated advantages $\hat{A}_{i}^{t} = m_{i}^{t} \cdot A_{i}^{t}$.
The scaling factor is determined by both the advantage sign and trajectory source:

\begin{small}
    \begin{equation}
        \label{eq:adv_mask}
        m_{i}^{t} :=
        \begin{cases}
            \alpha, & \quad \text{if } A_{i}^{t} > 0 \text{ and } \boldsymbol{\tau}_{i} \sim \pi_{r}                      \\[1ex]
            \beta,  & \quad \text{if } A_{i}^{t} < 0 \text{ and } \boldsymbol{\tau}_{i} \sim \pi_{n}                      \\
                    & \quad \text{and } \boldsymbol{\tau}_{i,t} \in \mathcal{S}_{\text{think}}                            \\[1ex]
            0,      & \quad \text{if } A_{i}^{t} > 0 \text{ and } \boldsymbol{\tau}_{i} \sim \pi_{n}                      \\
                    & \quad \text{and } \boldsymbol{\tau}_{i,t} \in \mathcal{S}_{\text{think}}                            \\
                    & \quad \text{and } \exists j \text{ s.t. } A_{j} > 0 \text{ and } \boldsymbol{\tau}_{j} \sim \pi_{r} \\[1ex]
            1,      & \quad \text{otherwise}
        \end{cases}
    \end{equation}
\end{small}
Equation~\eqref{eq:adv_mask} enables four distinct advantage operations, each of which targets a specific scenario:
\begin{itemize}[itemsep=0pt, leftmargin=*]
    \item \textbf{Enhancement ($m_{i}^{t} = \alpha > 1$):}
          Amplifies advantages for preferred trajectories from the rectified policy, strongly reinforcing thinking-free successful responses.

    \item \textbf{Suppression ($m_{i}^{t} = \beta > 1$):}
          Magnifies negative advantages for thinking tokens in unpreferred normal policy trajectories, actively discouraging overthinking behaviors that lead to incorrect responses.

    \item \textbf{Return-to-Zero ($m_{i}^{t} = 0$):}
          Eliminates advantages for thinking tokens in preferred normal policy trajectories when equivalent thinking-free solutions exist, removing redundant learning signals.

    \item \textbf{Identity ($m_{i}^{t} = 1$):}
          Preserves original advantages for all other tokens, maintaining standard GRPO behavior where differential treatment is unnecessary.
\end{itemize}
This design enables preferential reinforcement of concise preferred trajectories through enhancement while simultaneously suppressing the triggers (i.e., thinking tokens) of overthinking.
\looseness=-1

\subsubsection{Policy Shaping on Thinking Tokens} \label{sec: policy_shaping_on_thinking_tokens}
Our third innovation ensures consistent gradient flow for thinking token suppression by addressing GRPO's gradient clipping limitation.

\paragraph{GRPO Gradient Clipping Limitation.}
As established in Section~\ref{sec:policy_gradient_grpo}, GRPO's policy gradient for token $\boldsymbol{\tau}_{i,t}$ is zeroed when the importance ratio $r_{i}^{t}$ falls outside the clipping bounds $[1-\epsilon, 1+\epsilon]$. For thinking tokens in unpreferred trajectories (where $A_{i}^{t} < 0$), gradients are preserved only when $r_{i}^{t} > 1-\epsilon$, which requires:
$$\pi_{\theta}(\boldsymbol{\tau}_{i,t}|\boldsymbol{q},\boldsymbol{\tau}_{i,<t}) > (1-\epsilon) \cdot \pi_{\theta_{\text{old}}}(\boldsymbol{\tau}_{i,t}|\boldsymbol{q},\boldsymbol{\tau}_{i,<t})$$

However, this condition is difficult to satisfy due to thinking tokens' inherently high prediction probabilities. Section~\ref{sec:think_token_property} shows that the thinking token \textit{wait} being around $0.88$ in average probability, implying a high suppression threshold.
With the standard $\epsilon=0.2$, thinking tokens are suppressed only when
$\pi_{\theta}(\boldsymbol{\tau}_{i,t}|\boldsymbol{q},\boldsymbol{\tau}_{i,<t}) > (1-0.2) \times 0.88 = 0.704$.
Such high threshold causes critical suppression signals to be clipped away when we need them most.

\paragraph{Old Policy Calibration.}
To ensure consistent gradient flow for thinking token suppression, we calibrate the old policy probability for thinking tokens.
Inspired by LUFFY~\citep{yan2025learning}, we reshape the old policy probability as:

\begin{small}
    \begin{equation}
        \label{eq:old_policy_reshape}
        \hat{\pi}_{\theta_{\text{old}}} (\boldsymbol{\tau}_{i,t} | \boldsymbol{q}, \boldsymbol{\tau}_{i,<t} ) = \frac{\gamma}{1-\epsilon} \,, \quad \forall \boldsymbol{\tau}_{i,t} \in \mathcal{S}_{\text{think}}
    \end{equation}
\end{small}
where $\gamma$ is a small constant (we use $\gamma = 0.1$).
Then, the importance ratio is modified as:
\looseness=-1
\begin{small}
    \begin{align}
        \label{eq:importance_ratio_calibration}
        \hat{r}_{i}^{t} = \frac{\pi_{\theta} ({\boldsymbol{\tau}}_{i,t} |{q} ,{\boldsymbol{\tau}}_{i,<t} )}{\hat{\pi}_{\theta_{\text{old}}} ({\boldsymbol{\tau}}_{i,t} |{q} ,{\boldsymbol{\tau}}_{i,<t} )} = \frac{\pi_{\theta} ({\boldsymbol{\tau}}_{i,t} |{q} ,{\boldsymbol{\tau}}_{i,<t} )}{\gamma} \cdot (1-\epsilon)
    \end{align}
\end{small}

\paragraph{Guaranteed Gradient Flow.}
With this calibration, the clipping condition $\hat{r}_{i}^{t} > 1-\epsilon$ is satisfied whenever $\pi_{\theta}(\boldsymbol{\tau}_{i,t}|\boldsymbol{q},\boldsymbol{\tau}_{i,<t}) > \gamma$.
Since thinking tokens typically maintain probabilities well above $\gamma = 0.1$ during training, this ensures that suppression gradients for thinking tokens are rarely clipped.
Combined with our token-level advantage scaling, this strategy provides reliable and consistent learning signals for thinking token regulation throughout the training process.

\section{Experiments}\label{sec:experiments}
In this section, we discuss our experimental setup, detailed implementation and baselines.

\subsection{Setup}\label{sec:setup}
\textbf{Model.}
The training process is initialized with DeepSeek-R1-Distill-Qwen-1.5B, which is a popular LRM developed by~\citet{deepseekai2025deepseekr1}.
\looseness=-1

\noindent \textbf{Training Data.}
Inspired by recent works~\cite{qwen3,bae2025onlinedifficultyfilteringreasoning},
we prioritise the quality over quantity for RL training data, curating $1{,}000$ problems of medium difficulty
from the DAPO-MATH-17K dataset~\cite{yu2025dapoopensourcellmreinforcement}.
Specifically, we processed $28{,}000$ random DAPO-MATH-17K samples with the LRM, using $0.6$ temperature to generate $4$ responses per question with $16$K maximum length.
The final dataset comprised entries exhibiting an answer correctness rate between $0.25$ and $0.5$, coupled with an average response length exceeding 8192 tokens.
These selected data are of moderate difficulty and feature longer responses, making them prone to the thinking trap.
We name the curated dataset as DuPPO-1K.

\noindent \textbf{Validation Data.}
We utilise AIME24 as the validation set, following the setting of DAPO~\cite{yu2025dapoopensourcellmreinforcement}
and VAPO~\cite{yue2025vapoefficientreliablereinforcement}.
During inference on this validation set, the temperature is set to $0$.
Adhering to the principle of balancing performance and token usage optimally, we report the evaluation results for the following checkpoints: DeepSeek-R1-Distill-Qwen-1.5 with GRPO at 90-step, DeepSeek-R1-Distill-Qwen-1.5 with \method at $80$-step.

\begin{table*}[!htbp]
    \centering
    \caption{\textbf{Main results on six mathematical reasoning benchmarks.}
        \textbf{Score} represents the average performance, using Pass@1 for \textsc{MINERVA}, \textsc{MATH500}, and \textsc{OLYMPIADBENCH} (for short OLYMPIAD), and Avg@32 for \textsc{AMC}, \textsc{AIME24}, and \textsc{AIME25}.
        \textbf{Len} denotes the average response length.
        Higher \textbf{Score} values and lower \textbf{Len} values indicate better performance.
        Bold entries highlight the best result in each column.}
    \label{tab:main_results}
    \normalfont
    \resizebox{\textwidth}{!}{%
        \begin{tabular}{
                l
                cc cc cc cc cc cc
                >{\columncolor{LightYellow}}c
                >{\columncolor{LightYellow}}c
                >{\columncolor{LightYellow}}c
                >{\columncolor{LightYellow}}c}
            \toprule
            \multirow{2}{*}{\textbf{Model}}       &
            \multicolumn{2}{c}{\textbf{MATH500}}  &
            \multicolumn{2}{c}{\textbf{OLYMPIAD}} &
            \multicolumn{2}{c}{\textbf{MINERVA}}  &
            \multicolumn{2}{c}{\textbf{AIME24}}   &
            \multicolumn{2}{c}{\textbf{AMC}}      &
            \multicolumn{2}{c}{\textbf{AIME25}}   &
            \multicolumn{4}{c}{\textbf{Average}}                                                                                                                                                                                                                                                                                                                                                                  \\
            \cmidrule(lr){2-3}\cmidrule(lr){4-5}\cmidrule(lr){6-7}
            \cmidrule(lr){8-9}\cmidrule(lr){10-11}\cmidrule(lr){12-13}\cmidrule(lr){14-17}
                                                  & Score ($\uparrow$) & Len ($\downarrow$) & Score ($\uparrow$) & Len ($\downarrow$) & Score ($\uparrow$) & Len ($\downarrow$) & Score ($\uparrow$) & Len ($\downarrow$) & Score ($\uparrow$) & Len ($\downarrow$) & Score ($\uparrow$) & Len ($\downarrow$) & Score ($\uparrow$) & Len ($\downarrow$) & $\Delta$Score ($\uparrow$) & $\Delta$Len ($\downarrow$) \\
            \midrule
            \rowcolor{LightGray}\multicolumn{17}{l}{\textbf{DeepSeek-R1-Distill-Qwen-1.5B}}                                                                                                                                                                                                                                                                                                                       \\
            Base model                            & 79.2               & 3760               & 43.0               & 5992               & 29.3               & 4821               & 21.8               & 7410               & 55.4               & 5812               & 20.4               & 7277               & 43.9               & 6105               & --                         & --                         \\
            +\,NoThink                            & 72.3               & \textbf{2022}      & 40.5               & \textbf{4052}      & 24.4               & \textbf{1272}      & 21.9               & 6157               & 53.8               & 4153               & 18.5               & 6335               & 41.8               & 4502               & \(-2.1\)                   & \(-26.3\%\)                \\
            +\,ThinkTokenPenalty                  & 77.7               & 2547               & 43.8               & 4081               & 27.7               & 2392               & 20.1               & \textbf{5694}      & 57.2               & \textbf{3956}      & 18.8               & \textbf{5053}      & 44.0               & \textbf{4234}      & \(+0.1\)                   & \(-30.6\%\)                \\
            +\,GRPO                               & 81.4               & 3345               & 45.0               & 5614               & 30.1               & 4482               & 24.0               & 7169               & 58.7               & 5359               & 23.2               & 6956               & 46.6               & 5724               & \(+2.7\)                   & \(-6.2\%\)                 \\
            +\,\method                            & \textbf{82.7}      & 2830               & \textbf{47.4}      & 5033               & \textbf{30.6}      & 3585               & \textbf{25.0}      & 6795               & \textbf{60.9}      & 4722               & \textbf{21.7}      & 6499               & \textbf{47.9}      & 5162               & \(+4.0\)                   & \(-15.4\%\)                \\
            \bottomrule
        \end{tabular}}
\end{table*}

\noindent \textbf{Benchmarks and Metrics.}
To evaluate performance, we selected six popular math reasoning benchmarks: \textsc{AIME 2024}, \textsc{AIME 2025}, \textsc{AMC}~\cite{numina_math_datasets}, \textsc{Minerva}~\cite{lewkowycz2022solvingquantitativereasoningproblems}, \textsc{OlympiadBench}~\cite{he2024olympiadbenchchallengingbenchmarkpromoting}, and \textsc{MATH500}~\cite{math500}.
Given the comparatively smaller test sets of \textsc{AIME} 2024, \textsc{AIME} 2025, and \textsc{AMC}, we present results using Avg@32.
For \textsc{Minerva}, \textsc{OlympiadBench} and \textsc{MATH500}, Pass@1 is used.
The temperature is set to $0.6$, and the max response length is set to $8{,}192$ for testing.
To mitigate systematic errors caused by sampling randomness, all reported results represent the average of three inference runs.

\subsection{Baselines}\label{sec:baselines}
We compare \method against the base model (DeepSeek-R1-Distill-Qwen-1.5B) and two test-time efficiency baselines:
\begin{itemize}[nosep, leftmargin=12pt]
    \item \textbf{NoThink}~\cite{ma2025reasoning} bypasses internal reasoning by appending ``\textit{Okay, I think I have finished thinking.}</think>'' after ``<think>'' in prompts, forcing models to generate concise answers directly.
    \item \textbf{ThinkTokenPenalty}~\cite{underthinking-o1} applies the logit penalty to thinking tokens (e.g., \textit{however, alternatively}) during inference.
          We adopt an aggressive variant that penalizes all thinking tokens to prevent their sampling entirely.
          This technique aligns with sampling from the rectified policy $\pi_r$ described in Section~\ref{sec:duppo_2}.
\end{itemize}

\subsection{RL Practice}\label{sec:RL_Practice}
\subsubsection{Implementation Details}\label{sec:implement_details}
\paragraph{Training and Inference.}
We use VeRL~\cite{sheng2024hybridflow} to implement \method, where the training context size, batch size, and the learning rate are $8{,}192$, $128$ and $2e{-}6$.
The actor policy update batch size is $128$.
The total rollout number is set as $8$, including $N=4$ responses sampling from the normal policy $\pi_{n}$, and $M=4$ responses sampling from the rectified policy $\pi_{r}$.
We remove the KL loss term by setting $\beta=0$ and set the entropy loss coefficient to $0.01$.
In terms of the newly added hyperparameters of \method, we set the \textit{enhancement factor} $\alpha$ and the \textit{suppression factor} $\beta$ as $2$ for advantage scaling.
For the 1.5B model, we train them within $100$ steps.
All training experiments are conducted on a single 8xH800 node, and all inference experiments use four NVIDIA RTX 4090 GPUs.

\paragraph{Thinking Tokens.}
For thinking token identification, we categorize common thinking tokens in LRM responses into reflection tokens (\textit{wait, hmm, hold on, okay}) and thought transition tokens (\textit{alternatively, maybe, but, however}).
Our analysis in Section~\ref{sec:think_token_property} focuses on the \textit{wait} token as a representative case, while both \method and ThinkTokenPenalty penalize all identified thinking tokens to maximize token efficiency.

\subsubsection{Reward Design}\label{sec:reward_design}
We design a rule-based reward function that incorporates both correctness and formatting rewards, similar to the general RL training approach used in DeepSeek-R1~\cite{deepseekai2025deepseekr1}.
The reward function evaluates two components: answer correctness through Math-Verify parsing and response formatting quality.
Specifically, correct answers receive a base reward of $1.0$, with an additional $0.1$ bonus for well-formatted responses.
The reward function is formally defined as:
\begin{equation}
    \small
    \label{eq:reward_func}
    R(\boldsymbol{\tau} ,\boldsymbol{a})=\begin{cases}1.1,&\text{if } \  \texttt{is\_equivalent($\boldsymbol{a}, \boldsymbol{\tau}$)}\\ &\  \tau \  \text{is well-formatted}\\ 1.0,&\text{if } \  \texttt{is\_equivalent($\boldsymbol{a}, \boldsymbol{\tau}$),}\\ &\  \tau \  \text{is not well-formatted}\\ 0.1,&\text{if } \  \texttt{not\_equivalent($\boldsymbol{a}, \boldsymbol{\tau}$),}\\ &\tau \  \text{is well-formatted}\\ 0,&\text{if} \  \texttt{not\_equivalent($\boldsymbol{a}, \boldsymbol{\tau}$),} \ \\ &\tau \  \text{is not well-formatted}\end{cases}
\end{equation}
where \texttt{is\_equivalent($\boldsymbol{a}, \boldsymbol{\tau}$)} determines whether the ground truth answer $\boldsymbol{a}$ can be successfully parsed from the response trajectory $\boldsymbol{\tau}$.

\section{Results and Discussion}\label{sec:results_discussion} 
We conduct comprehensive experiments across six widely-adopted mathematical reasoning benchmarks to evaluate \method's effectiveness.
Our empirical analysis demonstrates consistent improvements in both performance and token efficiency, as summarized in Table~\ref{tab:main_results}.

\subsection{Comparision with Training-free Baselines}
We first establish the performance ceiling of training-free approaches.
While the prompt-based NoThink method exhibits the expected performance-efficiency trade-off ($26.3\%$ token reduction with a $2.1$-point accuracy penalty), \textbf{ThinkTokenPenalty} emerges as the superior training-free method, \textbf{achieving $30.6\%$ token reduction while preserving accuracy} on DeepSeek-R1-Distill-Qwen-1.5B.
This finding reveals that thinking tokens in the base model may contribute minimally to reasoning quality, primarily inflating response length.
Such result aligns with and extend the key insights from Section~\ref{sec:think_token_absence_maintain_perf} across a broader range of datasets, providing additional empirical validation that thinking tokens are largely dispensable for reasoning.

However, training-based optimization unlocks substantially greater potential.
Compared to ThinkTokenPenalty, \method outperforms by $3.9$ points while maintaining a significant token efficiency, demonstrating superior performance-efficiency trade-offs that training-free methods cannot achieve.

\begin{tcolorbox}[colback=green!10,colframe=green!50!black,title=\textbf{Takeaway}]
    Training-based optimization is essential: \method \textbf{substantially outperforms} all training-free approaches on performance.
\end{tcolorbox}

\subsection{Comparision with Base Models}
Our empirical evaluation on DeepSeek-R1-Distill-Qwen-1.5B reveals that \method achieves substantial improvements with minimal training overhead.
Specifically, the method requires only $80$ RL steps to deliver $4.0$ points average performance gain alongside $15.4\%$ token reduction.

\method achieves consistent performance improvements across benchmark complexities with substantial efficiency gains.
Simpler benchmarks like \textsc{MATH500} show pronounced benefits with $3.5$-point improvements and $24.7\%$ token savings.
On more challenging benchmarks such as \textsc{AIME24} and \textsc{AIME25}, performance gains remain consistent while token reduction becomes more conservative.
\textsc{AIME24}  improves from $21.8$ to $25.0$ with $8.3\%$ token reduction, and \textsc{AIME25} from $20.4$ to $21.7$ with $10.7\%$ fewer tokens.
These results indicate \method's adaptive optimization across problem complexities.
\begin{tcolorbox}[colback=green!10,colframe=green!50!black,title=\textbf{Takeaway}]
    \method consistently \textbf{improves both performance and token efficiency} over the base model with \textbf{minimal training cost}.
\end{tcolorbox}

\subsection{Comparision with GRPO}\label{sec:comparision_with_GRPO}
We conduct a controlled comparison against GRPO to isolate the contributions of our proposed innovations.
Standard GRPO training ($90$ steps) yields $2.7$ points average improvement with $6.2\%$ token reduction relative to the base model.

\method demonstrates clear algorithmic superiority.
With fewer training iterations ($80$ vs $90$ steps), our method achieves $1.3$ points higher accuracy than GRPO while consuming fewer reasoning tokens ($5{,}162$ vs $5{,}724$).
This demonstrates superior efficiency with reduced training iterations and lower token consumption during inference.

\begin{tcolorbox}[colback=green!10,colframe=green!50!black,title=\textbf{Takeaway}]
    \method outperforms GRPO in \textbf{performance}, \textbf{token efficiency}, and \textbf{training speed} simultaneously.
\end{tcolorbox}

\section{Conclusion}\label{sec:conclusion}
In this paper, we present \method, a novel reinforcement learning algorithm designed to address the overthinking problem in the $1.5$B LRM.
Our experimental results demonstrate that frequent use of thinking tokens is not necessarily essential for model performance improvement.
Through fine-grained control of thinking tokens, our method achieves superior balance between performance enhancement and token efficiency compared to baseline approaches, requiring only lightweight training on the base LRM across multiple mathematical benchmarks.
For future work, we plan to validate the reliability and robustness of our approach by extending experiments to larger model architectures and diverse domain benchmarks.

%% file: resources/appendix.tex
\onecolumn
{
    \hypersetup{linkcolor=black}
    \parskip=0em
    \renewcommand{\contentsname}{Contents}
    \tableofcontents
    \addtocontents{toc}{\protect\setcounter{tocdepth}{3}}
}

\newpage

%% file: paper.bbl
\begin{thebibliography}{51}
\providecommand{\natexlab}[1]{#1}

\bibitem[{Aggarwal and Welleck(2025)}]{aggarwal2025l1}
Pranjal Aggarwal and Sean Welleck. 2025.
\newblock L1: Controlling how long a reasoning model thinks with reinforcement learning.
\newblock \emph{arXiv preprint arXiv:2503.04697}.

\bibitem[{Bae et~al.(2025)Bae, Hong, Lee, Kim, Nam, and Kwak}]{bae2025onlinedifficultyfilteringreasoning}
Sanghwan Bae, Jiwoo Hong, Min~Young Lee, Hanbyul Kim, JeongYeon Nam, and Donghyun Kwak. 2025.
\newblock \href {https://arxiv.org/abs/2504.03380} {Online difficulty filtering for reasoning oriented reinforcement learning}.
\newblock \emph{Preprint}, arXiv:2504.03380.

\bibitem[{Chen et~al.(2025)Chen, Xu, Liang, He, Pang, Yu, Song, Liu, Zhou, Zhang, Wang, Tu, Mi, and Yu}]{chen2025do}
Xingyu Chen, Jiahao Xu, Tian Liang, Zhiwei He, Jianhui Pang, Dian Yu, Linfeng Song, Qiuzhi Liu, Mengfei Zhou, Zhuosheng Zhang, Rui Wang, Zhaopeng Tu, Haitao Mi, and Dong Yu. 2025.
\newblock \href {https://arxiv.org/abs/2412.21187} {Do not think that much for 2+3=? on the overthinking of o1-like llms}.
\newblock \emph{Preprint}, arXiv:2412.21187.

\bibitem[{Cui et~al.(2025)Cui, He, Zeng, Liu, Tang, Dai, Han, Luo, Huang, Li et~al.}]{cui2025stepwise}
Yingqian Cui, Pengfei He, Jingying Zeng, Hui Liu, Xianfeng Tang, Zhenwei Dai, Yan Han, Chen Luo, Jing Huang, Zhen Li, and 1 others. 2025.
\newblock Stepwise perplexity-guided refinement for efficient chain-of-thought reasoning in large language models.
\newblock \emph{arXiv preprint arXiv:2502.13260}.

\bibitem[{DeepSeek{-}AI(2025)}]{deepseekai2025deepseekr1}
DeepSeek{-}AI. 2025.
\newblock \href {https://doi.org/10.48550/ARXIV.2501.12948} {Deepseek-r1: Incentivizing reasoning capability in llms via reinforcement learning}.
\newblock \emph{CoRR}, abs/2501.12948.

\bibitem[{Fang et~al.(2025)Fang, Ma, and Wang}]{thinkless}
Gongfan Fang, Xinyin Ma, and Xinchao Wang. 2025.
\newblock \href {https://arxiv.org/abs/2505.13379} {Thinkless: Llm learns when to think}.
\newblock \emph{Preprint}, arXiv:2505.13379.

\bibitem[{Feng et~al.(2025)Feng, Fang, Ma, and Wang}]{feng2025efficientreasoningmodelssurvey}
Sicheng Feng, Gongfan Fang, Xinyin Ma, and Xinchao Wang. 2025.
\newblock \href {https://arxiv.org/abs/2504.10903} {Efficient reasoning models: A survey}.
\newblock \emph{Preprint}, arXiv:2504.10903.

\bibitem[{Fu et~al.(2024)Fu, Chen, Zhu, Fu, Dai, Qiao, and Zhang}]{fu2024efficiently}
Yichao Fu, Junda Chen, Siqi Zhu, Zheyu Fu, Zhongdongming Dai, Aurick Qiao, and Hao Zhang. 2024.
\newblock Efficiently serving llm reasoning programs with certaindex.
\newblock \emph{arXiv preprint arXiv:2412.20993}.

\bibitem[{He et~al.(2024)He, Luo, Bai, Hu, Thai, Shen, Hu, Han, Huang, Zhang, Liu, Qi, Liu, and Sun}]{he2024olympiadbenchchallengingbenchmarkpromoting}
Chaoqun He, Renjie Luo, Yuzhuo Bai, Shengding Hu, Zhen~Leng Thai, Junhao Shen, Jinyi Hu, Xu~Han, Yujie Huang, Yuxiang Zhang, Jie Liu, Lei Qi, Zhiyuan Liu, and Maosong Sun. 2024.
\newblock \href {https://arxiv.org/abs/2402.14008} {Olympiadbench: A challenging benchmark for promoting agi with olympiad-level bilingual multimodal scientific problems}.
\newblock \emph{Preprint}, arXiv:2402.14008.

\bibitem[{Hou et~al.(2025)Hou, Zhang, Ji, Liu, Qian, Andreas, and Chang}]{hou2025thinkprune}
Bairu Hou, Yang Zhang, Jiabao Ji, Yujian Liu, Kaizhi Qian, Jacob Andreas, and Shiyu Chang. 2025.
\newblock Thinkprune: Pruning long chain-of-thought of llms via reinforcement learning.
\newblock \emph{arXiv preprint arXiv:2504.01296}.

\bibitem[{{Jia LI} et~al.(2024){Jia LI}, {Edward Beeching}, {Lewis Tunstall}, {Ben Lipkin}, {Roman Soletskyi}, {Shengyi Costa Huang}, {Kashif Rasul}, {Longhui Yu}, {Albert Jiang}, {Ziju Shen}, {Zihan Qin}, {Bin Dong}, {Li Zhou}, {Yann Fleureau}, {Guillaume Lample}, and {Stanislas Polu}}]{numina_math_datasets}
{Jia LI}, {Edward Beeching}, {Lewis Tunstall}, {Ben Lipkin}, {Roman Soletskyi}, {Shengyi Costa Huang}, {Kashif Rasul}, {Longhui Yu}, {Albert Jiang}, {Ziju Shen}, {Zihan Qin}, {Bin Dong}, {Li Zhou}, {Yann Fleureau}, {Guillaume Lample}, and {Stanislas Polu}. 2024.
\newblock \href {https://github.com/project-numina/aimo-progress-prize/blob/main/report/numina_dataset.pdf} {Numinamath}.

\bibitem[{Kang et~al.(2025)Kang, Sun, Chen, and Zou}]{kang2025c3ot}
Yu~Kang, Xianghui Sun, Liangyu Chen, and Wei Zou. 2025.
\newblock C3ot: Generating shorter chain-of-thought without compromising effectiveness.
\newblock In \emph{Proceedings of the AAAI Conference on Artificial Intelligence}, volume~39, pages 24312--24320.

\bibitem[{Lewkowycz et~al.(2022)Lewkowycz, Andreassen, Dohan, Dyer, Michalewski, Ramasesh, Slone, Anil, Schlag, Gutman-Solo, Wu, Neyshabur, Gur-Ari, and Misra}]{lewkowycz2022solvingquantitativereasoningproblems}
Aitor Lewkowycz, Anders Andreassen, David Dohan, Ethan Dyer, Henryk Michalewski, Vinay Ramasesh, Ambrose Slone, Cem Anil, Imanol Schlag, Theo Gutman-Solo, Yuhuai Wu, Behnam Neyshabur, Guy Gur-Ari, and Vedant Misra. 2022.
\newblock \href {https://arxiv.org/abs/2206.14858} {Solving quantitative reasoning problems with language models}.
\newblock \emph{Preprint}, arXiv:2206.14858.

\bibitem[{Lightman et~al.(2024)Lightman, Kosaraju, Burda, Edwards, Baker, Lee, Leike, Schulman, Sutskever, and Cobbe}]{math500}
Hunter Lightman, Vineet Kosaraju, Yuri Burda, Harrison Edwards, Bowen Baker, Teddy Lee, Jan Leike, John Schulman, Ilya Sutskever, and Karl Cobbe. 2024.
\newblock \href {https://openreview.net/forum?id=v8L0pN6EOi} {Let's verify step by step}.
\newblock In \emph{The Twelfth International Conference on Learning Representations, {ICLR} 2024, Vienna, Austria, May 7-11, 2024}. OpenReview.net.

\bibitem[{Liu et~al.(2025)Liu, Chen, Li, Pang, Du, and Lin}]{liu2025there}
Zichen Liu, Changyu Chen, Wenjun Li, Tianyu Pang, Chao Du, and Min Lin. 2025.
\newblock There may not be aha moment in r1-zero-like training — a pilot study.
\newblock \url{https://oatllm.notion.site/oat-zero}.
\newblock Notion Blog.

\bibitem[{Luo et~al.(2025{\natexlab{a}})Luo, Shen, He, Wang, Liu, Li, Tan, Cao, and Tao}]{o1-pruner}
Haotian Luo, Li~Shen, Haiying He, Yibo Wang, Shiwei Liu, Wei Li, Naiqiang Tan, Xiaochun Cao, and Dacheng Tao. 2025{\natexlab{a}}.
\newblock \href {https://arxiv.org/abs/2501.12570} {O1-pruner: Length-harmonizing fine-tuning for o1-like reasoning pruning}.
\newblock \emph{Preprint}, arXiv:2501.12570.

\bibitem[{Luo et~al.(2025{\natexlab{b}})Luo, Shen, He, Wang, Liu, Li, Tan, Cao, and Tao}]{luo2025o1}
Haotian Luo, Li~Shen, Haiying He, Yibo Wang, Shiwei Liu, Wei Li, Naiqiang Tan, Xiaochun Cao, and Dacheng Tao. 2025{\natexlab{b}}.
\newblock O1-pruner: Length-harmonizing fine-tuning for o1-like reasoning pruning.
\newblock \emph{arXiv preprint arXiv:2501.12570}.

\bibitem[{Ma et~al.(2025{\natexlab{a}})Ma, He, Snell, Griggs, Min, and Zaharia}]{ma2025reasoning}
Wenjie Ma, Jingxuan He, Charlie Snell, Tyler Griggs, Sewon Min, and Matei Zaharia. 2025{\natexlab{a}}.
\newblock Reasoning models can be effective without thinking.
\newblock \emph{arXiv preprint arXiv:2504.09858}.

\bibitem[{Ma et~al.(2025{\natexlab{b}})Ma, Wan, Yu, Fang, and Wang}]{ma2025cot}
Xinyin Ma, Guangnian Wan, Runpeng Yu, Gongfan Fang, and Xinchao Wang. 2025{\natexlab{b}}.
\newblock Cot-valve: Length-compressible chain-of-thought tuning.
\newblock \emph{arXiv preprint arXiv:2502.09601}.

\bibitem[{Muennighoff et~al.(2025{\natexlab{a}})Muennighoff, Yang, Shi, Li, Fei{-}Fei, Hajishirzi, Zettlemoyer, Liang, Cand{\`{e}}s, and Hashimoto}]{muennighoff2025s1}
Niklas Muennighoff, Zitong Yang, Weijia Shi, Xiang~Lisa Li, Li~Fei{-}Fei, Hannaneh Hajishirzi, Luke Zettlemoyer, Percy Liang, Emmanuel~J. Cand{\`{e}}s, and Tatsunori Hashimoto. 2025{\natexlab{a}}.
\newblock \href {https://doi.org/10.48550/ARXIV.2501.19393} {s1: Simple test-time scaling}.
\newblock \emph{CoRR}, abs/2501.19393.

\bibitem[{Muennighoff et~al.(2025{\natexlab{b}})Muennighoff, Yang, Shi, Li, Fei-Fei, Hajishirzi, Zettlemoyer, Liang, Candès, and Hashimoto}]{muennighoff2025s1simpletesttimescaling}
Niklas Muennighoff, Zitong Yang, Weijia Shi, Xiang~Lisa Li, Li~Fei-Fei, Hannaneh Hajishirzi, Luke Zettlemoyer, Percy Liang, Emmanuel Candès, and Tatsunori Hashimoto. 2025{\natexlab{b}}.
\newblock \href {https://arxiv.org/abs/2501.19393} {s1: Simple test-time scaling}.
\newblock \emph{Preprint}, arXiv:2501.19393.

\bibitem[{Munkhbat et~al.(2025)Munkhbat, Ho, Kim, Yang, Kim, and Yun}]{munkhbat2025self}
Tergel Munkhbat, Namgyu Ho, Seo~Hyun Kim, Yongjin Yang, Yujin Kim, and Se-Young Yun. 2025.
\newblock Self-training elicits concise reasoning in large language models.
\newblock \emph{arXiv preprint arXiv:2502.20122}.

\bibitem[{{Open-R1-Team}(2025)}]{huggingface2025mini}
{Open-R1-Team}. 2025.
\newblock \href {https://huggingface.co/blog/open-r1/mini-r1-contdown-game} {Mini r1 countdown game}.
\newblock Blog post.

\bibitem[{{OpenAI}(2024)}]{openai2024reasoning}
{OpenAI}. 2024.
\newblock Learning to reason with llms.
\newblock \url{https://openai.com/index/learning-to-reason-with-llms/}.
\newblock Accessed: 2025-05-07.

\bibitem[{Qian et~al.(2025)Qian, Liu, Wen, Bai, Liu, and Shao}]{qian2025demystifying}
Chen Qian, Dongrui Liu, Haochen Wen, Zhen Bai, Yong Liu, and Jing Shao. 2025.
\newblock \href {https://arxiv.org/abs/2506.02867} {Demystifying reasoning dynamics with mutual information: Thinking tokens are information peaks in llm reasoning}.
\newblock \emph{Preprint}, arXiv:2506.02867.

\bibitem[{Qu et~al.(2025)Qu, Yang, Setlur, Tunstall, Beeching, Salakhutdinov, and Kumar}]{qu2025optimizing}
Yuxiao Qu, Matthew~YR Yang, Amrith Setlur, Lewis Tunstall, Edward~Emanuel Beeching, Ruslan Salakhutdinov, and Aviral Kumar. 2025.
\newblock Optimizing test-time compute via meta reinforcement fine-tuning.
\newblock \emph{arXiv preprint arXiv:2503.07572}.

\bibitem[{{Qwen Team}(2025)}]{qwen2025qwq}
{Qwen Team}. 2025.
\newblock Qwq-32b-preview.
\newblock \url{https://qwenlm.github.io/blog/qwq-32b-preview/}.
\newblock Accessed: 15 March 2025.

\bibitem[{Schulman et~al.(2017)Schulman, Wolski, Dhariwal, Radford, and Klimov}]{ppo}
John Schulman, Filip Wolski, Prafulla Dhariwal, Alec Radford, and Oleg Klimov. 2017.
\newblock \href {https://api.semanticscholar.org/CorpusID:28695052} {Proximal policy optimization algorithms}.
\newblock \emph{ArXiv}, abs/1707.06347.

\bibitem[{Seed et~al.(2025)Seed, :, Chen, Fan, Liu, Liu, Lin, Wang, Wang, Wei, Xu, Yuan, Yue, Yan, Yu, Zuo, Zhang, Zhu, An, Bai, Bao, Bin, Chen, Chen, Chen, Chen, Chen, Chen, Chen, Chen, Chen, Chen, Chen, Chen, Chi, Dai, Dai, Dai, Dou, Du, Du, Duan, Dun, Fan, Feng, Feng, Feng, Fu, Fu, Fu, Ge, Guo, Han, Han, Hao, Hao, He, He, He, Heng, Hong, Hou, Hu, Hu, Hu, Hua, Huang, Huang, Huang, Huang, Huang, Huang, Jia, Jia, Jia, Jiang, Jiang, Jiang, Jiang, Jiang, Jiao, Jin, Jin, Lai, Li, Li, Li, Li, Li, Wan, Wang, Li, Li, Li, Li, Li, Li, Li, Li, Liang, Liang, Lin, Lin, Lin, Liu, Liu, Liu, Liu, Liu, Liu, Liu, Liu, Liu, Liu, Liu, Liu, Liu, Liu, Liu, Liu, Long, Lou, Lou, Luo, Luo, Lv, Lv, Ma, Ma, Ma, Ma, Ma, Ma, Ma, Mao, Min, Nan, Ning, Ou, Pan, Pang, Peng, Peng, Qian, Qian, Qiao, Qu, Ren, Ren, Shan, Shen, Shen, Shen, Sheng, Shi, Shi, Shi, Cao, Song, Song, Su, Sun, Sun, Sun, Wan, Wang, Wang, Wang, Wang, Wang, Wang, Wang, Wang, Wang, Wang, Wang, Wang, Wang, Wang, Wang, Wang, Wang, Wei, Wei, Wei, Wei, Wu, Wu, Wu, Wu, Wu, Wu, Wu, Wu, Wu, Xi, Xia, Xian, Xiang, Xiang, Xiao, Xiao, Xiao, Xiao, Xin, Xin, Xiong, Xu, Xu, Xu, Xu, Xu, Xu, Xu, Xu, Yan, Yan, Yang, Yang, Yang, Yang, Yang, Yang, Yang, Yang, Yang, Yao, Yi, Yin, Yin, Ying, Yu, Yu, Yu, Yu, Yu, Yuan, Yuan, Zeng, Zhan, Zhang, Zhang, Zhang, Zhang, Zhang, Zhang, Zhang, Zhang, Zhang, Zhang, Zhang, Zhang, Zhang, Zhang, Zhang, Zhang, Zhang, Zheng, Zheng, Zheng, Zheng, Zheng, Zhi, Zhong, Zhong, Zhong, Zhong, Zhou, Zhou, Zhou, Zhu, Zhu, Zhu, and Zuo}]{seed2025seed15thinking}
ByteDance Seed, :, Jiaze Chen, Tiantian Fan, Xin Liu, Lingjun Liu, Zhiqi Lin, Mingxuan Wang, Chengyi Wang, Xiangpeng Wei, Wenyuan Xu, Yufeng Yuan, Yu~Yue, Lin Yan, Qiying Yu, Xiaochen Zuo, Chi Zhang, Ruofei Zhu, Zhecheng An, and 255 others. 2025.
\newblock \href {https://arxiv.org/abs/2504.13914} {Seed1.5-thinking: Advancing superb reasoning models with reinforcement learning}.
\newblock \emph{Preprint}, arXiv:2504.13914.

\bibitem[{Shen et~al.(2025)Shen, Zhang, Huang, Shi, Zhang, Yan, Wang, Wang, Liu, and Lian}]{shen2025dast}
Yi~Shen, Jian Zhang, Jieyun Huang, Shuming Shi, Wenjing Zhang, Jiangze Yan, Ning Wang, Kai Wang, Zhaoxiang Liu, and Shiguo Lian. 2025.
\newblock Dast: Difficulty-adaptive slow-thinking for large reasoning models.
\newblock \emph{arXiv preprint arXiv:2503.04472}.

\bibitem[{Sheng et~al.(2024)Sheng, Zhang, Ye, Wu, Zhang, Zhang, Peng, Lin, and Wu}]{sheng2024hybridflow}
Guangming Sheng, Chi Zhang, Zilingfeng Ye, Xibin Wu, Wang Zhang, Ru~Zhang, Yanghua Peng, Haibin Lin, and Chuan Wu. 2024.
\newblock Hybridflow: A flexible and efficient rlhf framework.
\newblock \emph{arXiv preprint arXiv: 2409.19256}.

\bibitem[{Su and Cardie(2025)}]{thinkingfastrightbalancing}
Jinyan Su and Claire Cardie. 2025.
\newblock \href {https://arxiv.org/abs/2505.18298} {Thinking fast and right: Balancing accuracy and reasoning length with adaptive rewards}.
\newblock \emph{Preprint}, arXiv:2505.18298.

\bibitem[{Sui et~al.(2025)Sui, Chuang, Wang, Zhang, Zhang, Yuan, Liu, Wen, Zhong, Chen, and Hu}]{sui2025stopoverthinkingsurveyefficient}
Yang Sui, Yu{-}Neng Chuang, Guanchu Wang, Jiamu Zhang, Tianyi Zhang, Jiayi Yuan, Hongyi Liu, Andrew Wen, Shaochen Zhong, Hanjie Chen, and Xia~Ben Hu. 2025.
\newblock \href {https://doi.org/10.48550/ARXIV.2503.16419} {Stop overthinking: {A} survey on efficient reasoning for large language models}.
\newblock \emph{CoRR}, abs/2503.16419.

\bibitem[{Team et~al.(2025)Team, Du, Gao, Xing, Jiang, Chen, Li, Xiao, Du, Liao et~al.}]{team2025kimi}
Kimi Team, Angang Du, Bofei Gao, Bowei Xing, Changjiu Jiang, Cheng Chen, Cheng Li, Chenjun Xiao, Chenzhuang Du, Chonghua Liao, and 1 others. 2025.
\newblock Kimi k1.5: Scaling reinforcement learning with llms.
\newblock \emph{arXiv preprint arXiv:2501.12599}.

\bibitem[{Tu et~al.(2025)Tu, Lin, Zhang, Tian, Li, Lan, and Zhao}]{tu2025learning}
Songjun Tu, Jiahao Lin, Qichao Zhang, Xiangyu Tian, Linjing Li, Xiangyuan Lan, and Dongbin Zhao. 2025.
\newblock Learning when to think: Shaping adaptive reasoning in r1-style models via multi-stage rl.
\newblock \emph{arXiv preprint arXiv:2505.10832}.

\bibitem[{Wang et~al.(2025{\natexlab{a}})Wang, Yu, Gao, Zheng, Liu, Lu, Dang, Chen, Yang, Zhang, Liu, Yang, Zhao, Yue, Song, Yu, Huang, and Lin}]{wang2025beyond}
Shenzhi Wang, Le~Yu, Chang Gao, Chujie Zheng, Shixuan Liu, Rui Lu, Kai Dang, Xionghui Chen, Jianxin Yang, Zhenru Zhang, Yuqiong Liu, An~Yang, Andrew Zhao, Yang Yue, Shiji Song, Bowen Yu, Gao Huang, and Junyang Lin. 2025{\natexlab{a}}.
\newblock \href {https://arxiv.org/abs/2506.01939} {Beyond the 80/20 rule: High-entropy minority tokens drive effective reinforcement learning for llm reasoning}.
\newblock \emph{Preprint}, arXiv:2506.01939.

\bibitem[{Wang et~al.(2025{\natexlab{b}})Wang, Liu, Xu, Liang, Chen, He, Song, Yu, Li, Zhang, Wang, Tu, Mi, and Yu}]{underthinking-o1}
Yue Wang, Qiuzhi Liu, Jiahao Xu, Tian Liang, Xingyu Chen, Zhiwei He, Linfeng Song, Dian Yu, Juntao Li, Zhuosheng Zhang, Rui Wang, Zhaopeng Tu, Haitao Mi, and Dong Yu. 2025{\natexlab{b}}.
\newblock \href {https://arxiv.org/abs/2501.18585} {Thoughts are all over the place: On the underthinking of o1-like llms}.
\newblock \emph{Preprint}, arXiv:2501.18585.

\bibitem[{Wang et~al.(2025{\natexlab{c}})Wang, Liu, Xu, Liang, Chen, He, Song, Yu, Li, Zhang et~al.}]{wang2025thoughts}
Yue Wang, Qiuzhi Liu, Jiahao Xu, Tian Liang, Xingyu Chen, Zhiwei He, Linfeng Song, Dian Yu, Juntao Li, Zhuosheng Zhang, and 1 others. 2025{\natexlab{c}}.
\newblock Thoughts are all over the place: On the underthinking of o1-like llms.
\newblock \emph{arXiv preprint arXiv:2501.18585}.

\bibitem[{Xia et~al.(2025)Xia, Li, Leong, Wang, and Li}]{xia2025tokenskip}
Heming Xia, Yongqi Li, Chak~Tou Leong, Wenjie Wang, and Wenjie Li. 2025.
\newblock Tokenskip: Controllable chain-of-thought compression in llms.
\newblock \emph{arXiv preprint arXiv:2502.12067}.

\bibitem[{Yan et~al.(2025)Yan, Li, Hu, Wang, Cui, Qu, Cheng, and Zhang}]{yan2025learning}
Jianhao Yan, Yafu Li, Zican Hu, Zhi Wang, Ganqu Cui, Xiaoye Qu, Yu~Cheng, and Yue Zhang. 2025.
\newblock Learning to reason under off-policy guidance.
\newblock \emph{arXiv preprint arXiv:2504.14945}.

\bibitem[{Yang et~al.(2025{\natexlab{a}})Yang, Li, Yang, Zhang, Hui, Zheng, Yu, Gao, Huang, Lv, Zheng, Liu, Zhou, Huang, Hu, Ge, Wei, Lin, Tang, Yang, Tu, Zhang, Yang, Yang, Zhou, Zhou, Lin, Dang, Bao, Yang, Yu, Deng, Li, Xue, Li, Zhang, Wang, Zhu, Men, Gao, Liu, Luo, Li, Tang, Yin, Ren, Wang, Zhang, Ren, Fan, Su, Zhang, Zhang, Wan, Liu, Wang, Cui, Zhang, Zhou, and Qiu}]{qwen3}
An~Yang, Anfeng Li, Baosong Yang, Beichen Zhang, Binyuan Hui, Bo~Zheng, Bowen Yu, Chang Gao, Chengen Huang, Chenxu Lv, Chujie Zheng, Dayiheng Liu, Fan Zhou, Fei Huang, Feng Hu, Hao Ge, Haoran Wei, Huan Lin, Jialong Tang, and 41 others. 2025{\natexlab{a}}.
\newblock \href {https://arxiv.org/abs/2505.09388} {Qwen3 technical report}.
\newblock \emph{Preprint}, arXiv:2505.09388.

\bibitem[{Yang et~al.(2025{\natexlab{b}})Yang, Si, Duan, Zhu, Zhu, Li, Lin, Cao, and Wang}]{yang2025dynamic}
Chenxu Yang, Qingyi Si, Yongjie Duan, Zheliang Zhu, Chenyu Zhu, Qiaowei Li, Zheng Lin, Li~Cao, and Weiping Wang. 2025{\natexlab{b}}.
\newblock Dynamic early exit in reasoning models.
\newblock \emph{arXiv preprint arXiv:2504.15895}.

\bibitem[{Yang et~al.(2025{\natexlab{c}})Yang, Lin, and Yu}]{yang2025think}
Junjie Yang, Ke~Lin, and Xing Yu. 2025{\natexlab{c}}.
\newblock Think when you need: Self-adaptive chain-of-thought learning.
\newblock \emph{arXiv preprint arXiv:2504.03234}.

\bibitem[{Yang et~al.(2025{\natexlab{d}})Yang, Wu, Chen, Xiao, Yang, Wong, and Wang}]{yang2025understanding}
Shu Yang, Junchao Wu, Xin Chen, Yunze Xiao, Xinyi Yang, Derek~F. Wong, and Di~Wang. 2025{\natexlab{d}}.
\newblock \href {https://arxiv.org/abs/2504.02956} {Understanding aha moments: from external observations to internal mechanisms}.
\newblock \emph{Preprint}, arXiv:2504.02956.

\bibitem[{Yeo et~al.(2025)Yeo, Tong, Niu, Neubig, and Yue}]{yeo2025demystifying}
Edward Yeo, Yuxuan Tong, Morry Niu, Graham Neubig, and Xiang Yue. 2025.
\newblock Demystifying long chain-of-thought reasoning in llms.
\newblock \emph{arXiv preprint arXiv:2502.03373}.

\bibitem[{Yu et~al.(2024)Yu, Xu, Weston, and Kulikov}]{yu2024distilling}
Ping Yu, Jing Xu, Jason Weston, and Ilia Kulikov. 2024.
\newblock Distilling system 2 into system 1.
\newblock \emph{arXiv preprint arXiv:2407.06023}.

\bibitem[{Yu et~al.(2025)Yu, Zhang, Zhu, Yuan, Zuo, Yue, Dai, Fan, Liu, Liu, Liu, Lin, Lin, Ma, Sheng, Tong, Zhang, Zhang, Zhang, Zhu, Zhu, Chen, Chen, Wang, Yu, Song, Wei, Zhou, Liu, Ma, Zhang, Yan, Qiao, Wu, and Wang}]{yu2025dapoopensourcellmreinforcement}
Qiying Yu, Zheng Zhang, Ruofei Zhu, Yufeng Yuan, Xiaochen Zuo, Yu~Yue, Weinan Dai, Tiantian Fan, Gaohong Liu, Lingjun Liu, Xin Liu, Haibin Lin, Zhiqi Lin, Bole Ma, Guangming Sheng, Yuxuan Tong, Chi Zhang, Mofan Zhang, Wang Zhang, and 16 others. 2025.
\newblock \href {https://arxiv.org/abs/2503.14476} {Dapo: An open-source llm reinforcement learning system at scale}.
\newblock \emph{Preprint}, arXiv:2503.14476.

\bibitem[{Yue et~al.(2025)Yue, Yuan, Yu, Zuo, Zhu, Xu, Chen, Wang, Fan, Du, Wei, Yu, Liu, Liu, Liu, Lin, Lin, Ma, Zhang, Zhang, Zhang, Zhu, Zhang, Liu, Wang, Wu, and Yan}]{yue2025vapoefficientreliablereinforcement}
Yu~Yue, Yufeng Yuan, Qiying Yu, Xiaochen Zuo, Ruofei Zhu, Wenyuan Xu, Jiaze Chen, Chengyi Wang, TianTian Fan, Zhengyin Du, Xiangpeng Wei, Xiangyu Yu, Gaohong Liu, Juncai Liu, Lingjun Liu, Haibin Lin, Zhiqi Lin, Bole Ma, Chi Zhang, and 8 others. 2025.
\newblock \href {https://arxiv.org/abs/2504.05118} {Vapo: Efficient and reliable reinforcement learning for advanced reasoning tasks}.
\newblock \emph{Preprint}, arXiv:2504.05118.

\bibitem[{Zeng et~al.(2025)Zeng, Huang, Liu, Liu, He, Ma, and He}]{zeng2025simplerlzooinvestigatingtamingzero}
Weihao Zeng, Yuzhen Huang, Qian Liu, Wei Liu, Keqing He, Zejun Ma, and Junxian He. 2025.
\newblock \href {https://arxiv.org/abs/2503.18892} {Simplerl-zoo: Investigating and taming zero reinforcement learning for open base models in the wild}.
\newblock \emph{Preprint}, arXiv:2503.18892.

\bibitem[{Zhang et~al.(2025)Zhang, Lin, Hou, Feng, and Li}]{zhang2025adaptthink}
Jiajie Zhang, Nianyi Lin, Lei Hou, Ling Feng, and Juanzi Li. 2025.
\newblock Adaptthink: Reasoning models can learn when to think.
\newblock \emph{arXiv preprint arXiv:2505.13417}.

\bibitem[{Zhou et~al.(2025)Zhou, Li, Wang, Cheng, Zhou, and Hsieh}]{zhou2025r1zero}
Hengguang Zhou, Xirui Li, Ruochen Wang, Minhao Cheng, Tianyi Zhou, and Cho-Jui Hsieh. 2025.
\newblock \href {https://arxiv.org/abs/2503.05132} {R1-zero's "aha moment" in visual reasoning on a 2b non-sft model}.
\newblock \emph{Preprint}, arXiv:2503.05132.

\end{thebibliography}
